\documentclass[journal]{IEEEtai}

\usepackage[colorlinks,urlcolor=blue,linkcolor=blue,citecolor=blue]{hyperref}

\usepackage{color,array}

\usepackage{graphicx}
\usepackage{cite}
\usepackage[numbers,sort&compress]{natbib}
\usepackage{amsmath}
\usepackage{multirow}
\usepackage{subfigure}
\usepackage{booktabs}
\usepackage{authblk}

\begin{document}

\title{Time and Frequency Network for Human Action Detection in Videos} 

\author[1]{Changhai Li}
\author[1*]{Huawei Chen}
\author[2]{Jingqing Lu}
\author[1]{Yang Huang}
\author[1]{Yingying Liu}
\affil[1]{School of Aeronautics and Astronautics, University Of Electronic Science And Technology Of China}
\affil[2]{School of Life Science and Technology, University Of Electronic Science And Technology Of China}
\affil[*]{Corresponding author: chenhuawei@uestc.edu.cn}

\renewcommand*{\Affilfont}{\small\it} 
\renewcommand\Authands{ and } 
\date{} 

\maketitle

\begin{abstract}
Currently, spatiotemporal features are embraced by most deep learning approaches for human action detection in videos, however, they neglect the important features in frequency domain. In this work, we propose an end-to-end network that considers the time and frequency features simultaneously, named TFNet. TFNet holds two branches, one is time branch formed of three-dimensional convolutional neural network(3D-CNN), which takes the image sequence as input to extract time features; and the other is frequency branch, extracting frequency features through two-dimensional convolutional neural network(2D-CNN) from DCT coefficients. Finally, to obtain the action patterns, these two features are deeply fused under the attention mechanism. Experimental results on the JHMDB51-21 and UCF101-24 datasets demonstrate that our approach achieves remarkable performance for frame-mAP.
\end{abstract}


\section{Introduction}

\IEEEPARstart{I}{n} recent years, the attention on human action detection has been growing due to its extensive application in realistic scenes, such as video surveillance, human-computer interaction, and device control.  An action executed by the human body is a continuous process in time, using an image sequence with a  three-dimensional shape to manifest human action provides more distinguishing features than a single static frame. If only a single static frame is used, although the position of an object can be located exactly, it is challenging to recognize the action category. In Fig. 1, we can derive running-related motion information from long consecutive frames, such as arm swinging and leg stretching, which promotes the probability of this action to be accurately recognized as running.

\begin{figure}
	\centerline{\includegraphics[width=18.5pc]{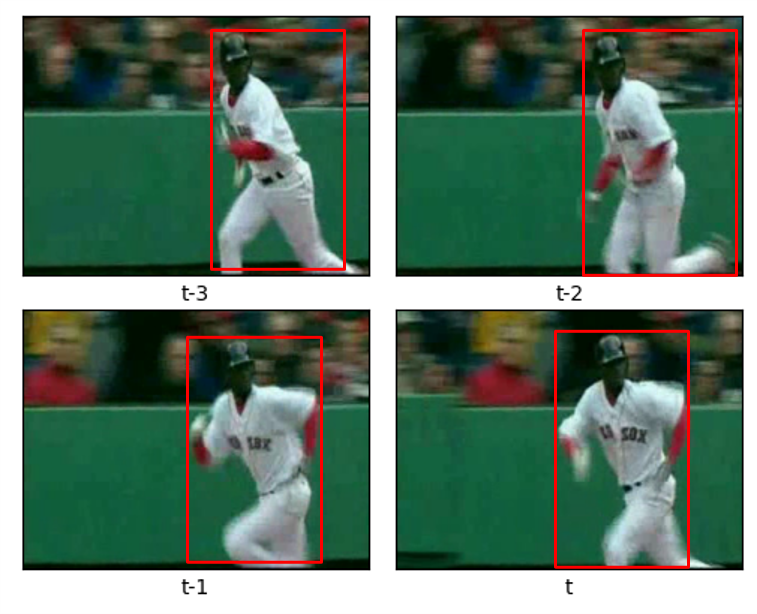}}
	\caption{``run'' action: 4 consecutive frames from JHMDB51-21 dataset}
\end{figure}

\begin{figure*}
	\centerline{\includegraphics[width=40pc]{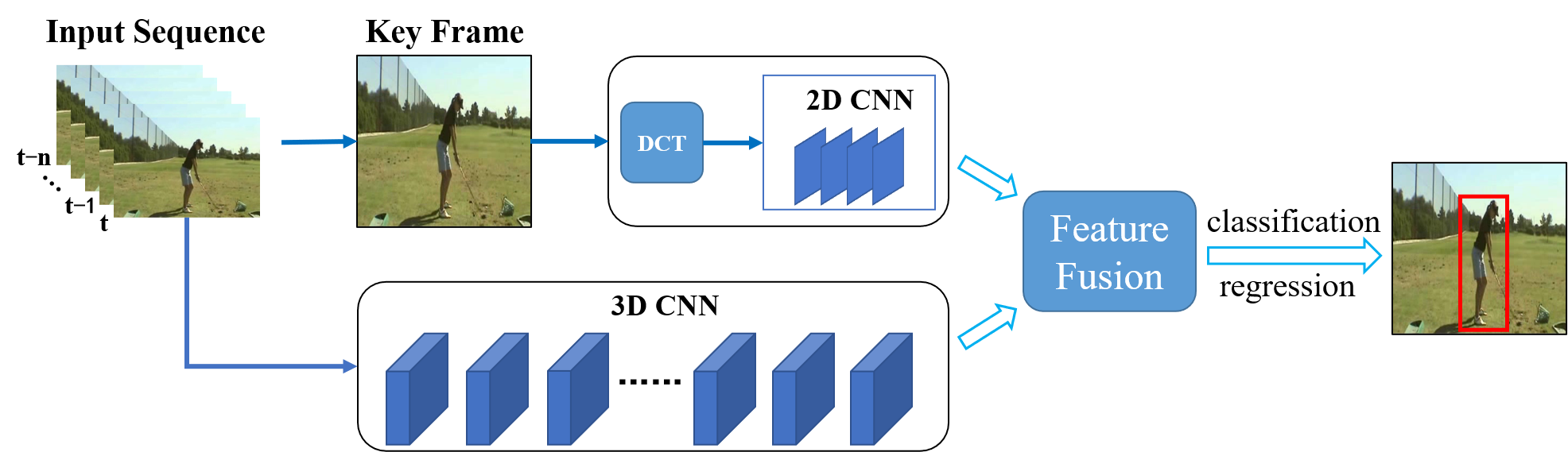}}
	\caption{The TFNet architecture. An input sequence and corresponding keyframe are fed to a 3D-CNN and 2D-CNN to produce time and frequency features, respectively. Then both two features are fused deeply for classification and bounding boxes regression.}
\end{figure*}

Convolutional neural network (CNN),  a widely used feature extraction technique, has an excellent performance in various computer vision tasks (e.g. video classification, object detection, and image segmentation). Also, most approaches are built upon CNN in human action detection. In 2014, Simonyan {\em et al.}\cite{simonyan2014twostream} proposed a two-stream CNN framework, they thought that the spatial component in the video contains the static features of the background and the object, while the temporal component provides the feature representation of the optical flow between frames. Accordingly, they built a two-stream model to extract spatial and temporal feature information separately. Although this approach promotes the application and development of CNN for human action detection, there are still several disadvantages that exist. In this case, the computation of optical flow and fine-tuning parameters is carried out in stages, which complicates the entire training process. Besides, it takes a long time to calculate the optical flow, which cannot be applied in some scenes with high real-time demands. Based on 2D-CNN, Shuiwang {\em et al.}\cite{Shuiwang2013} appended the time dimension and proposed a 3D-CNN model for human action detection first time. They extracted motion-related features through a three-dimensional convolutional kernel. Experimental results show the progressiveness in this way and avoid the above drawbacks caused by the optical flow. Subsequently, extensive convolutional networks based on 3D-CNN have been continuously proposed. The more typical one is the C3D network proposed by Du Tran {\em et al.}\cite{Tran2015ICCV} in 2015. From a series of experiments, they found that: 1) the classification performance of 3D-CNN exceeds 2D-CNN in videos; 2) the small convolutional kernel of size $3 \times 3 \times 3$  used uniformly in all convolutional layers achieves better performance ; 3) C3D network can be applied to other computer vision tasks easily. Although 3D-CNN extracts spatiotemporal features efficiently, a noteworthy issue is that 3D-CNN inevitably consumes more computing resources than 2D-CNN, and requires larger memory capacity. Hence, based on the existed 3D-CNN and skip connection, Qiu {\em et al.}\cite{Qiu2017ICCV} employed $1 \times 3 \times 3$ spatial kernel and $3 \times 1 \times 1$ temporal kernel to approximately replace the $3 \times 3 \times 3$ kernel and proposed P3D residual network with fewer parameters in 2017. 

Most approaches\cite{Kalogeiton2017ICCV,Jianchao2020,Xiaojiang2016,Song2019CVPR,Hou2017ICCV,kopuklu2019yowo} for human action detection based on CNN, only consider the spatiotemporal features, while frequency features are neglected. Nevertheless, the image spectrums provide a priori distribution information in the frequency domain, which facilitates the learning of convolutional networks. However, two different images may have the same or similar frequency spectrums, which will raise the error detection. Consequently, additional information must be linked to determine the final detection result. Several remarkable works\cite{Gueguen2018,Xu2020CVPR,Ulicny2017OnUC} have successfully applied frequency-domain-learning approaches to image classification and instance segmentation, but not for human action detection. Inspired by the preceding works, in this paper, we propose an end-to-end Time and Frequency Network (TFNet) as shown in Fig. 2, to the best of our knowledge, which is the first application of frequency-domain-learning approaches on human action detection. TFNet holds two branches in the time and frequency domain. The time branch is implemented by a 3D-CNN to extract time features. The frequency branch firstly converts the keyframe to frequency coefficients by DCT transformation, then 2D-CNN is employed to derive features from frequency coefficients. Later, A feature fusion algorithm based on the attention mechanism is used to deeply fuse the time-frequency domain features extracted from both two branches.

Contributions of our work are summarized as follows: 1) we propose an end-to-end single stage network in the time-frequency domain, called TFNet, which can efficiently extract time-frequency features for human action detection; 2) we also develop an attention mechanism-based fusing algorithm to make features mingle deeply from time-frequency domain.

\section{Methodology}
In this section, we will present TFNet architecture in detail. We first describe the implementation of DCT transformation, then backbones in both two branches are introduced, finally, we also describe the theory of feature fusion algorithm.

\subsection{DCT Transformation}
Most of the natural images we see are in the RGB color space which puts brightness, hue, and saturation together to express a color effect of the display, but the components of R, G, and B are highly correlated, if a certain component changes to a particular extent, then this color is likely to be modified, which is not conducive to DCT transformation.\cite{DCT1974} The YCbCr color space is usually adopted in DCT transformation due to it separates brightness and hue components in different color channels and its encoding mode saves storage space. According to the above analysis, before doing DCT transformation, the keyframe needs to be converted from RGB to YCbCr color space, and then the following Equation 1 is executed to transform separately for each color channel. In Equation 1, $f(x,y)$ is an image block, $M$ and $N$ denote the height and width of the image block, respectively, $x$ and $y$ are the indexes of location response to $x\in{[0, M-1]}$ and $y\in[0, N -1]$. The value of factor $\alpha(k)$ is shown in Equation 2,  this factor does not affect the properties of the DCT transformation, multiplying by the factor is to satisfy the orthogonality of the DCT transform matrix.

\begin{equation}
\begin{split}
    &F(u,v)=\alpha (u)\alpha (v)  \\ 
    &\sum_{x=0}^{M-1}\sum_{y=0}^{N-1}f(x,y)cos(\frac{(2x+1)u\pi}{2M})cos(\frac{(2y+1)v\pi}{2N})
\end{split}
\end{equation}

\begin{equation}
    \alpha (k)=\begin{cases}
    \sqrt{\frac{1}{N}}& \text{ if } k=0 \\ 
    \sqrt{\frac{2}{N}}& \text{ if } k=1,2,...,N-1 
    \end{cases}
\end{equation}

\begin{figure*}
	\centerline{\includegraphics[width=40pc]{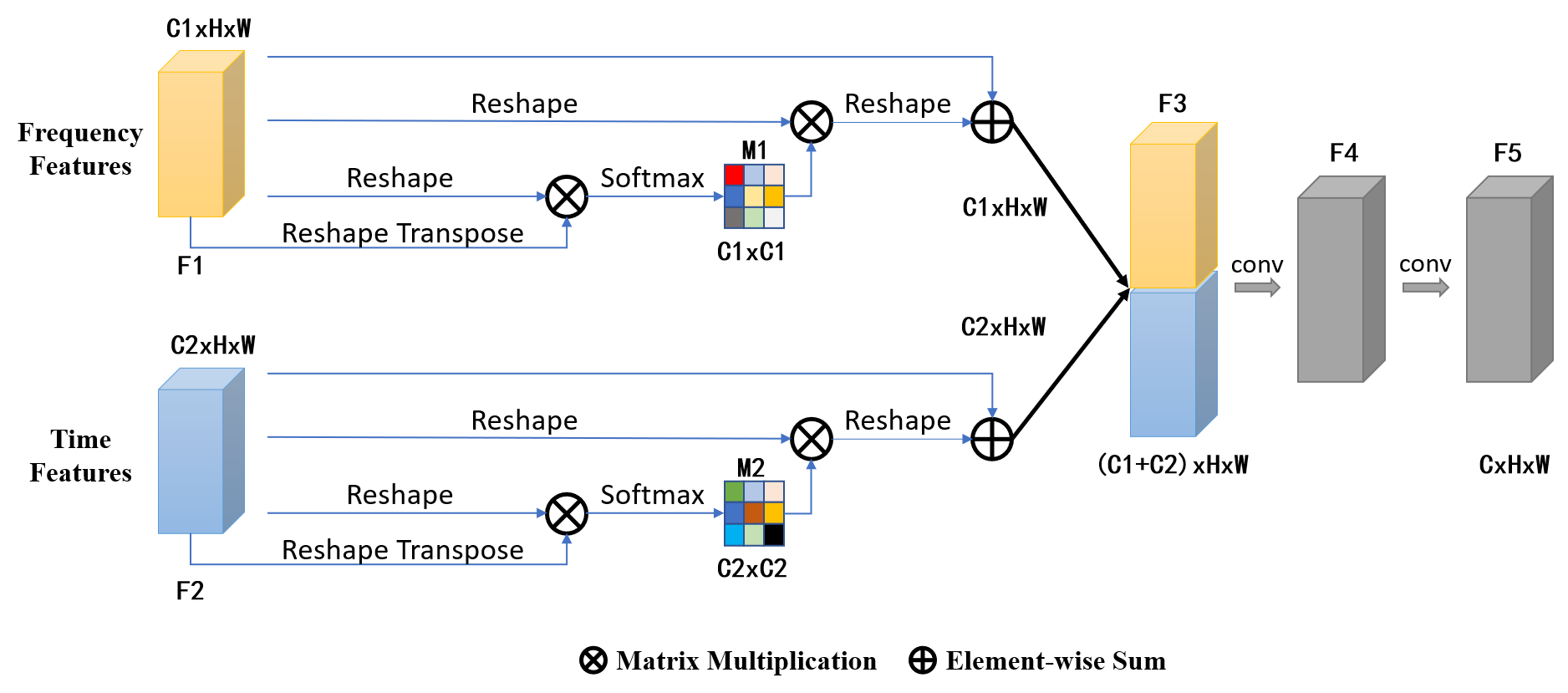}}
	\caption{Feature fusion algorithm base on attention mechanism for aggregating time and frequency feature maps coming from 3D-CNN and 2D-CNN.}
\end{figure*}

According to the JPEG\cite{JPEG1992} compression algorithm, we first divide each channel in the YCbCr color space into $8 \times 8$ pixel blocks with non-overlapping. The boundaries of width and height are expanded with 0 so that the size can be divisible by 8. After the division, the DCT transformation is performed on these $8 \times 8$ pixel blocks in turn to get the frequency coefficients. The coefficients with the same frequency are grouped into the same channel, so each color component has 64 channels with different frequencies.\cite{Gueguen2018} We don't have to use all 64 channels, under the balance of performance, we can choose the appropriate number of input channels. In this work, we control the number of input channels $C_{input}$ through the hyperparameter $\lambda$. As shown in Equation 3, the value $\lambda$ ranges from 0 to 1. The ${\rm round}(x)$ function rounds the multiplication result, and the $\max(x,1)$ function takes the maximum value between $x$ and 1 to ensure that the number of selected channels is at least 1.

\begin{equation}
    C_{input}=\max({\rm round}(\lambda \times 64), 1) \quad (0\leq \lambda \leq 1)
\end{equation}

\subsection{Frequency Branch}
To learn the frequency feature from DCT coefficients, we send DCT coefficients into 2D-CNN of the frequency branch. Since the Darknet-19\cite{Redmon2017CVPR} network has a great balance between accuracy and speed, for overall performance, we select the Darknet-19 network as the backbone to extract frequency features. Besides, to match the shape of the input DCT coefficients, the first 4 layers and the last 1 layer of convolution in the Darknet-19 backbone network, including the pooling layers and ReLU functions following them, are excluded, the configuration of the other layers remains consistent. As a result, the input shape of this branch is $C_f \times 56 \times 56$, where $C_f$ is the number of channels of DCT coefficients, which can be set during the training process, and the output shape of frequency features is fixed at $425 \times 7 \times 7$.

\subsection{Time Branch}
As described in Section 1, contextual information in time dimension is extremely crucial for human action detection. In this section, we exploit a 3D-CNN network as the main module to extract features in this time branch.  Meanwhile, the 3D-ResNeXt-101\cite{Hara2018CVPR}  model exhibits exceptional performance on Kinetics\cite{carreira2019short}, the largest human action detection dataset currently. There is no doubt that we choose the 3D-ResNeXt-101 model as the backbone network. The input of this branch is the original image sequence with the shape $C \times D \times H \times W$, corresponding to the channel, depth, height, and width respectively. The number of channels is 3, the total counts of color components R, G, B in one image. Depth is equal to the number of frames in the image sequence, which will finally be pooled to 1 so that the outputs of the current branch can match the frequency branch's results in dimension, which is beneficial for the subsequent feature fusion process in the channel dimension. The height and width of the input image have a fixed size of $224 \times 224$ that transform from other sizes in preprocess stage. Specifically, the size of input images is [3, D, 224, 224], at the final output of this branch,  we obtain the time feature maps with size [2048, 7, 7].

\subsection{Feature Fusion}
After extracting features from the time and frequency branches, we fuse these two features with an attention mechanism\cite{Fu2019CVPR} as shown in Figure 3.  We first obtain attention features instead of fusing directly from the original. Next, several operations are employed to the attention features. The overall principle is as follows: assuming that frequency features denote as $F1 \in R^{C1 \times H \times W}$, we merge height and width channels as one to create $F \in R^{C1 \times L}$, where $L=H \times W$, then a matrix multiplication between F and its transpose $F^T \in R^{L \times C1}$ is executed to obtain $G \in R^{C1 \times C1}$, later, the attention matrix $M1 \in R^{C1 \times C1}$ which indicates the feature correlations across channels is produced after the ${\rm softmax}(x)$ function, the above process is shown in Equation 4. From the inside, elements of M1 are calculated using Equation 5. 

\begin{equation}
    \begin{cases}
      G=F\cdot F^T &  \\ 
      M1={\rm softmax}(G) & 
    \end{cases}
\end{equation}

\begin{equation}
    M1_{ij}=\frac{{\rm exp}(G_{ij})}{\sum_{j=1}^{C1}{\rm exp}(G_{ij})}
\end{equation}

\begin{table*}[]
\setlength{\abovecaptionskip}{0pt} 
\setlength{\belowcaptionskip}{3pt}
\caption{We report the results of comparative experiments on JHMDB51-21. The AP  and mAP of all the classes are computed. ID:Action lable, TF:TFNet, T:Time-only Model}
\label{table}
\centering
\setlength{\tabcolsep}{5pt}
\tablefont%
\renewcommand\arraystretch{2}
\begin{tabular}{|c|c|c|c|c|c|c|c|c|c|c|c|c|c|c|c|}
\hline
\multicolumn{2}{|c|}{\textbf{Dataset}}             & \multicolumn{12}{c|}{\textbf{AP}}                                                             & \multicolumn{2}{c|}{\textbf{mAP}}                    \\ \hline
\multirow{6}{*}{\textbf{\rotatebox{90}{JHMDB51-21}}} & \textbf{ID} & 1     & 2     & 3     & 4     & 5     & 6     & 7     & 8     & 9     & 10    & 11    & 12    & \multirow{3}{*}{\textbf{TF}} & \multirow{3}{*}{67.3} \\ \cline{2-14}
                                     & \textbf{TF} & 93.33 & 59.73 & 96.7  & 87.26 & 94.77 & 30.01 & 50.01 & 68.36 & 97.76 & 99.89 & 91.69 & 30.93 &                              &                       \\ \cline{2-14}
                                     & \textbf{T}  & 96.27 & 1.41  & 75.41 & 34.36 & 37.82 & 11.5  & 23.78 & 39.49 & 84.43 & 81.55 & 74.12 & 18.4  &                              &                       \\ \cline{2-16} 
                                     & \textbf{ID} & 13    & 14    & 15    & 16    & 17    & 18    & 19    & 20    & 21    &       &       &       & \multirow{3}{*}{\textbf{T}}  & \multirow{3}{*}{40.9} \\ \cline{2-14}
                                     & \textbf{TF} & 41.42 & 97.28 & 80.52 & 24.99 & 27.76 & 92.46 & 36.94 & 61.8  & 49.73 &       &       &       &                              &                       \\ \cline{2-14}
                                     & \textbf{T}  & 3.62  & 78.1  & 66.07 & 8.76  & 27.28 & 0     & 12.57 & 26.5  & 58.06 &       &       &       &                              &                       \\ \hline
\end{tabular}
\end{table*}

Significantly,  we need to add attention to the original frequency features with the attention matrix $M1$ by performing Equation 6. In this equation, producing $F' \in R^{C1 \times L}$ with matrix multiplication between the attention matrix $M1$ and deformed frequency feature F, and then transform $F'$ back to $F'' \in R^{C1 \times H \times W}$ with its original shape to conduct an affine transformation likely operation, as shown in Equation 7, where the coefficient factor $\alpha$ is a learnable weight parameter.

\begin{equation}
    F'=M1 \cdot F
\end{equation}

\begin{equation}
    F1'=\alpha F'' + F1
\end{equation}

We get attention feature $F1' \in R^{C1 \times H \times W}$ of the frequency branch from the weighting result in Equation 7, for the time attention feature $F2' \in R^{C2 \times H \times W}$, just follow the steps as like $F1'$. After that, we begin to merge these two attention features by concatenating them in the channel dimension simply since we have made these two features equal in size of height and width in advance.  In this way, we have got relatively shallow fusing feature $F3 \in R^{(C1+C2) \times H \times W}$,  which cannot be effectively incorporated in both two features. Therefore, two convolutional layers are appended at the end of F3. Finally, we can obtain the deeply fusing features, i.e. $F5 \in R^{C \times H \times W}$. Note that TFNet is end-to-end, the outputs of F5 are utilized to perform classification and bounding box regression operations, which is consistent with YOLOv2\cite{Redmon2017CVPR}. Each grid point is responsible for predicting 5 bounding boxes, and the information of each bounding box includes 4 position coordinates, 1 confidence, and the probabilities corresponding to each category. Hence the size of the output channel $C$ is $C=(4+1+N_{cls}) \times 5$, where $N_{cls}$ is the number of categories in the dataset.

\section{Experiments}
To verify the effectiveness of the proposed TFNet, we perform a series of experiments on the JHMDB51-21\cite{Jhuang2013ICCV} and UCF101-24\cite{soomro2012ucf101} datasets. In this section, we first introduce the datasets, implementation details, and evaluation metrics, then analyze the performance of TFNet through ablation experiments, and compare TFNet with previous approaches at the end.

\subsection{Dataset}
The JHMDB51-21 dataset is a sub-dataset separated from the HMDB51\cite{Kuehne2011} dataset. This sub-dataset contains 21 actions with a single person, and each action category has 36 to 55 video clips. Besides, each video clip is composed of ranging from 15 to 40 frames. and there are a total of 31838 frames in the whole dataset. 

UCF101-24 dataset contains 24 actions with categories and bounding boxes. The actions in the same category are divided into 25 groups based on background or actors, each group includes 4 to 7 video clips, which means that similar features can be extracted from the same video groups.

\subsection{Implementation Details and Evaluation Metrics}
In this work, we use the PyTorch deep learning framework to implement the TFNet model and train on the machine with two GPUs of RTX 2080 Ti. In the training process, a mini-batch Stochastic Gradient Descent (SGD) algorithm is adopted to optimize the parameters. Due to the limitation of GPU memory, the mini-batch is set to 12 and the momentum is set to 0.9. Meanwhile, the number of frames is 16, the resolution of frames is uniformly converted to $224 \times 224$, and the stride for selecting frames in the entire clip is randomly selected between 1 and 2, but fix to 1 for verification or testing.  We utilize the pre-trained weights from the ImageNet\cite{Deng2009} and Kinetics datasets to initialize the parameters in the frequency and time branch, respectively. In the feature fusion module,  the Xavier normal distribution is utilized to initialize the parameters of each layer. Further, adjusting the learning rate to finetune all the parameters. The initial value of the learning rate is 0.0001,  it decreases to half of itself every 10K iterations, and stops after 25 epochs. We have also taken data augmentation for each frame in the image sequence, including color jitter, random cropping, rescaling, and horizontal flipping.

For the detection task, a detection is True Positive means that the IoU(Intersection over Union) between the detected and ground truth bounding box is greater than the threshold, and the detected category is also correct. Here, we utilize mAP (mean Average Precision), the most commonly used evaluation metrics,  to measure the performance of TFNet, the threshold of IoU is set to 0.5. mAP reveals only the average detection ability of the model,  to further explore TFNet's potential, we also compare the classification and localization accuracy of the complete dataset.

\begin{figure*}
\centering
\caption{Saliency maps in time branch. The first row is original images, the second and third rows are the saliency maps of TFNet and the Time-only Model in order.}

\subfigure[1-8 saliency maps]{
    \includegraphics[width=40pc]{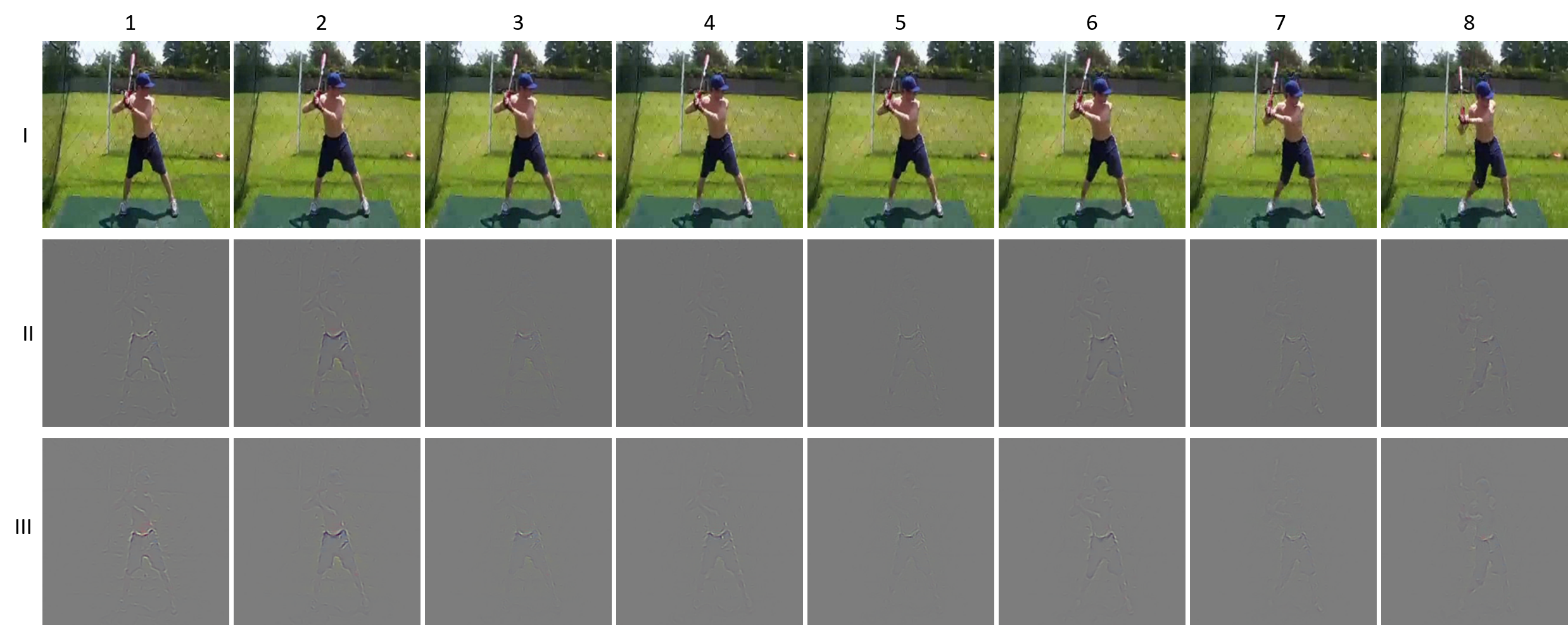}
}
\subfigure[9-16 saliency maps]{
    \includegraphics[width=40pc]{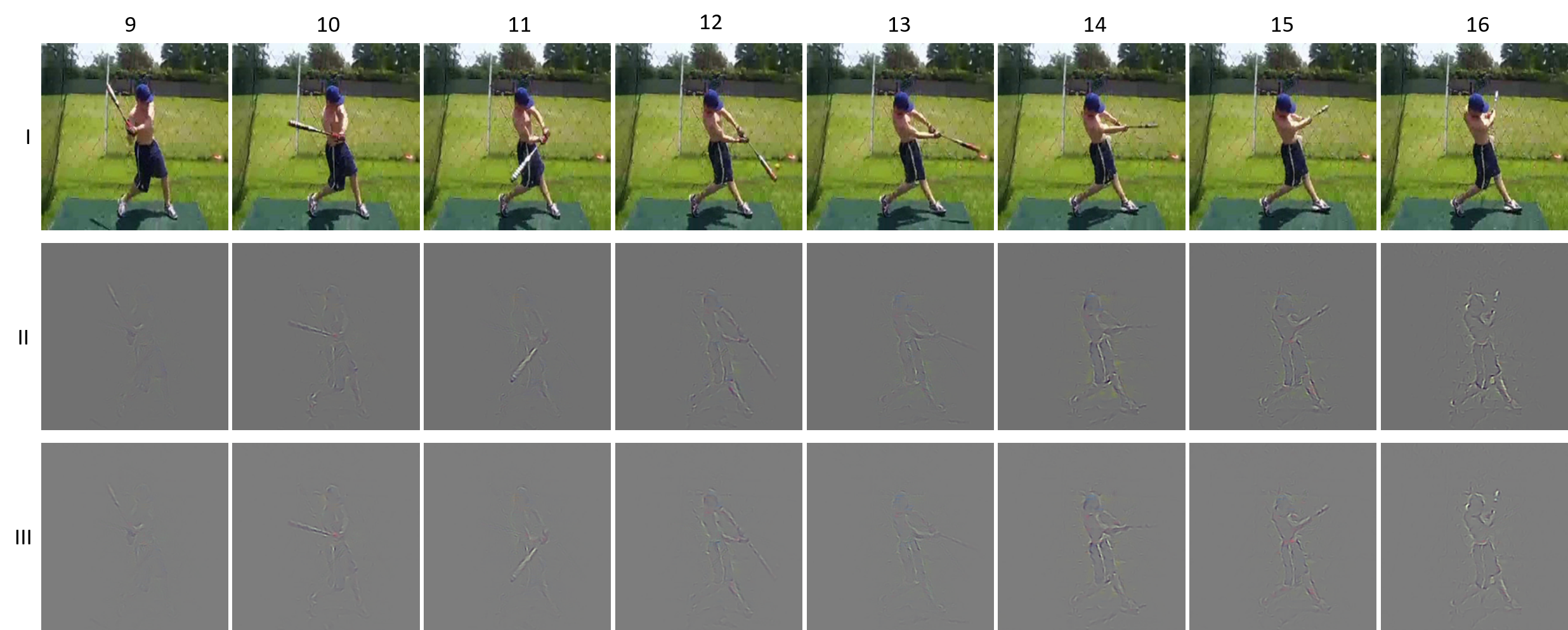}
}
\end{figure*}

\subsection{Performance Analysis}
We train another network model that only contains the time branch(Time-only Model) on the JHMDB51-21 dataset to test the effectiveness of TFNet in the same experimental conditions, the AP of each class and mAP(i.e. the average of all classes' APs) are calculated to be compared. The results are reported in Table 1, where the ID row represents the action classes with numbers in order, and TF and T rows record the results of  TFNet and Time-only Model respectively. From the results, we can see that TFNet has an extraordinary performance compare to the Time-only Model, in which the mAP of TFNet is overwhelmingly increased from 40.9\% of the Time-only Model to 67.3\%, a total gain of 26.4\%. Specifically, the APs of all action classes are increased except that class 1(brush\_hair) and class 21(wave). Among these improved action classes, actions of ID 2(catch), ID 13(shoot\_ball), and ID 18(swing\_baseball) have been affected significantly. The AP of ``catch'' has improved from 1.41\% to 59.73\%, and from 3.62\% to 41.42\% for ``shoot\_ball'', particularly, for ``swing\_baseball'', AP is zero in the Time-only Model, which means that this model cannot recognize ``swing\_baseball'', while TFNet has a superior detecting ability for this action, achieving an amazing score of 92.46\%.

To explore the factors that affect the performance of TFNet, we utilize the Guided-Backpropagation\cite{springenberg2015striving} approach to obtain saliency feature maps of the input. Guided-Backpropagation adds guidance based on ordinary backpropagation, avoiding the backpropagation of negative gradients, which will weaken the activated features. Therefore, the Guided-Backpropagation approach can find out activated features that are maximized.

\begin{figure*}
\centerline{\includegraphics[width=40pc]{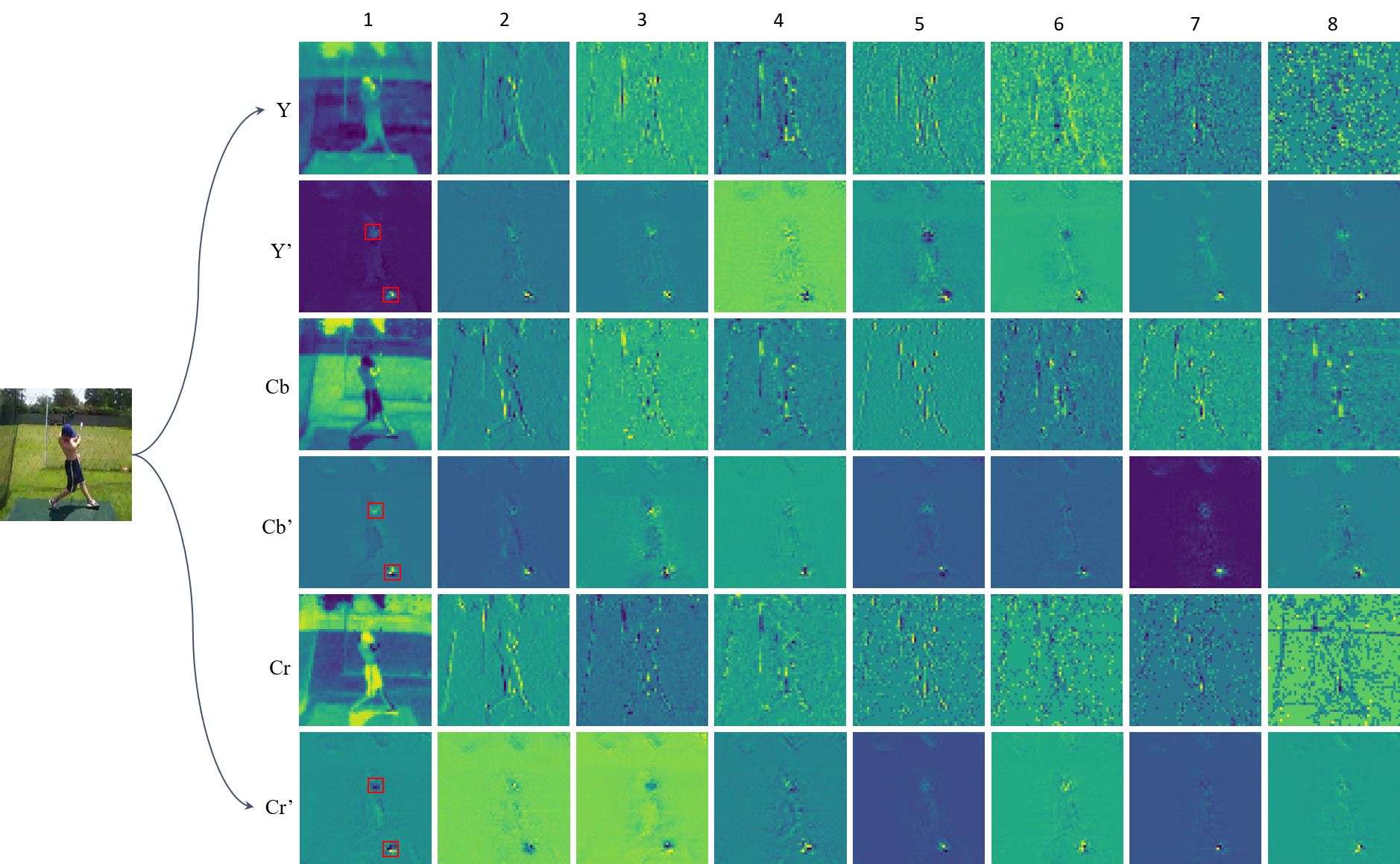}}
\caption{Saliency maps in frequency branch. The Y, Cb, and Cr rows are the first 8 channels selected from the DCT coefficients of the 3 color components, rows Y', Cb', and Cr' correspond to the saliency feature maps of rows Y, Cb, and Cr, respectively.}
\end{figure*}

Since the detected result of ``swing\_baseball'' has a huge distinction between these two models, we take saliency maps of this action as an instance to illustrate, see Fig. 4. The total 16 frames are exhibited in a and b of Fig. 4 of the first row separately,  the second and third rows are the saliency maps of TFNet and the Time-only Model in order. From the subfigures,  it can be found that the maximize activated features,  including the waist, thighs, bat, and body contour, are quite similar. However, these features are too ordinary to express the action of ``swing\_baseball'' uniquely, leading a garble with others, e.g. golf. That is the reason why the Time-only Model responses so poor for this action. In this case, maybe time branches of both two models are stuck in local optimal points during finetuning the parameters, and training losses are at a lower level, even tend to converge, where it is hard to finetune local parameters to promote overall performance.

The above problems can be solved by integrating other information, which is frequency features for TFNet, see Fig. 5. The Y, Cb, and Cr rows are the first 8 channels selected from the DCT coefficients of the 3 color components, and the frequencies of these 8 channels are the lowest among all channels. Since for higher frequency channels, the structure of the visualized images is difficult to identify through our eyes. Rows Y', Cb', and Cr' correspond to the saliency feature maps of rows Y, Cb, and Cr, respectively. Two places in these feature maps are highlighted. It is obvious from the DC channel that the features of arms and feet are aggregated. Although it cannot be intuitively distinguished from the higher frequency channels, there should be a certain pattern that has been learned by TFNet, which further improves the accuracy of detection.

\begin{table}[]
	\setlength{\abovecaptionskip}{0pt} 
	\setlength{\belowcaptionskip}{0pt}
	\centering
	\setlength{\tabcolsep}{7pt}
	\caption{The compare results between TFNet and other State-of-the-art approaches on JHMDB51-21 dataset. (IoU=0.5, frames=16)}
	\label{table}
	\tablefont%
	\renewcommand\arraystretch{2}
	\begin{tabular}{@{}ccc@{}}
		\toprule
		\textbf{Method} & \textbf{cls-accuracy} & \textbf{mAP}  \\ 
		\midrule
		MR-TS R-CNN\cite{Xiaojiang2016}     & 71.08                 & 58.5          \\
		T-CNN\cite{Hou2017ICCV}             & 67.2                  & 61.3          \\
		TACNet\cite{Song2019CVPR}           & -                     & 65.5          \\
		ACT\cite{Kalogeiton2017ICCV}        & 61.7                  & 65.7          \\ 
		\textbf{TFNet}                      & \textbf{81.5}         & \textbf{67.3} \\ 
		\bottomrule
	\end{tabular}
\end{table}

\begin{table}[]
	\setlength{\abovecaptionskip}{0pt} 
	\setlength{\belowcaptionskip}{0pt}
	\centering
	\setlength{\tabcolsep}{7pt}
	\caption{The compare results between TFNet and other State-of-the-art approaches on UCF101-24 dataset. (IoU=0.5, frames=16)}
	\label{table}
	\tablefont%
	\renewcommand\arraystretch{2}
	\begin{tabular}{@{}ccc@{}}
		\toprule
		\textbf{Method}        & \textbf{cls-accuracy} & \textbf{mAP}   \\
		\midrule
		Singh G, Saha S, {\em et al}.\cite{saha2016deep} & -             & 35.86    \\
		T-CNN\cite{Hou2017ICCV}                         & 94.4          & 41.37    \\
		MR-TS R-CNN\cite{Xiaojiang2016}                 & -             & 65.73    \\
		ACT\cite{Kalogeiton2017ICCV}                    & 84.7          & 67.1     \\
		\textbf{TFNet}                                  & \textbf{94.9} & \textbf{78.29} \\ \bottomrule
	\end{tabular}
\end{table}

To further analyze the influence of the frequency branch on the performance of TFNet, we perform a series of comparative experiments with different channels of DCT coefficients. The experimental results are shown in Fig. 6. From the figure, we can see that the number of channels has a great impact on the accuracy of classification and localization. As the number of channels increases, the accuracy of classification and localization cannot increase steadily, but we can achieve the best results if we utilize all 64 channels.

\subsection{State-of-the-art Comparison}
We first compare our TFNet with other state-of-the-art approaches on the JHMDB51-21 dataset. We report results in Table 2 under the condition of IoU equals 0.5 and the number of frames is 16. TFNet with 64 DCT channels surpasses all other state-of-the-art approaches in terms of cls-accuracy and mAP. In particular, TFNet far exceeds ACT\cite{Kalogeiton2017ICCV} in cls-accuracy by nearly 20\%, but the gap in mAP is tiny, only 1.6\%. TACNet\cite{Song2019CVPR} does not calculate cls-accuracy, while its mAP is still 1.8\% lower than our TFNet. Compared with T-CNN\cite{Hou2017ICCV} and ACT, MR-TS R-CNN\cite{Xiaojiang2016} has significantly raised cls-accuracy, but mAP has been decreased, which can not keep a balance between cls-accuracy and mAP. Nevertheless, our TFNet achieves a higher cls-accuracy while still maintaining an impressing performance of mAP.

We list the comparing results of the UCF101-24 dataset in Table 3. Under the same conditions, TFNet also produced a higher accuracy, with mAP reaches 78.29\%, which is 11.19\% higher than ACT. Meanwhile, in terms of cls-accuracy, it is 10.2\% higher than the ACT as well.

\begin{figure}
\centerline{\includegraphics[width=21pc]{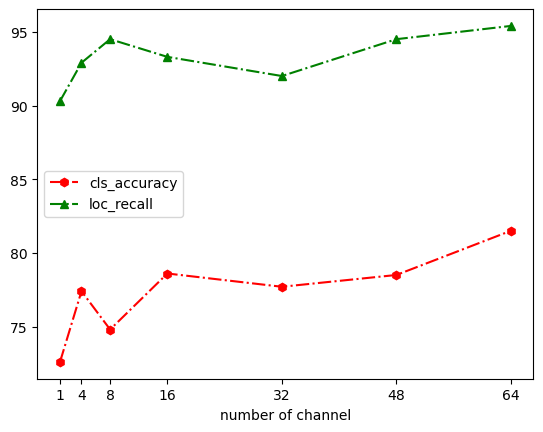}}
\caption{The experimental results on the JHMDB51-21 dataset. We test the impact of different numbers of DCT channels from 1 to 64}
\end{figure}

\section{Conclusion}
In this paper, we propose an end-to-end Time and Frequency Network for human action detection, i.e. TFNet. TFNet utilizes 2D-CNN and 3D-CNN to extract the time and frequency features of human action,  and fuse both two features with an attention mechanism to obtain detecting patterns. We conduct a series of comparative experiments on the JHMDB51-21 and UCF101-24 datasets. The experimental results demonstrate that our TFNet outperforms other state-of-the-art approaches, and also prove that it is feasible to detect actions by combining frequency features. However, there are still some human actions that have not been improved obviously. In future work, we would continue to explore larger datasets with more complicated actions.

\bibliographystyle{IEEEtran}
\bibliography{ref} 

\end{document}